%% file: neurips_2024.tex
\documentclass{article}

\usepackage[final, nonatbib]{neurips_2024}

\usepackage[utf8]{inputenc} 
\usepackage[T1]{fontenc}    
\usepackage{hyperref}       
\usepackage{url}            
\usepackage{booktabs}       
\usepackage{amsfonts}       
\usepackage{nicefrac}       
\usepackage{microtype}      
\usepackage{xcolor}         
\usepackage[sorting=none]{biblatex}
\usepackage{graphicx}
\usepackage{amsmath}
\usepackage{amssymb}

\usepackage{subcaption}
\usepackage{caption}

\newcommand{\mb}{\mathbf}

\title{CONCLAD: COntinuous Novel CLAss Detector}

\author{Amanda Rios\\
Intel\\
{\tt\small amanda.rios@intel.com}
\And
Ibrahima Ndiour\\
Intel\\
{\tt\small ibrahima.j.ndiour@intel.com}
\And
Parual Datta\\
Intel\\
{\tt\small parual.datta@intel.com}
\And
Omesh Tickoo\\
Intel\\
{\tt\small omesh.tickoo@intel.com}
\And
Nilesh Ahuja\\
Intel\\
{\tt\small nilesh.ahuja@intel.com}
}

\begin{document}

\maketitle

\vspace{-7mm}
\begin{abstract}
   \vspace{-3mm}
  In the field of continual learning, relying on so-called oracles for novelty detection is commonplace albeit unrealistic. This paper introduces CONCLAD ("COntinuous Novel CLAss Detector"), a comprehensive solution to the under-explored problem of continual novel class detection in post-deployment data. At each new task, our approach employs an iterative uncertainty estimation algorithm to differentiate between known and novel class(es) samples, and to further discriminate between the different novel classes themselves. Samples predicted to be from a novel class with high-confidence are automatically pseudo-labeled and used to update our model. Simultaneously, a tiny supervision budget is used to iteratively query ambiguous novel class predictions, which are also used during update. Evaluation across multiple datasets, ablations and experimental settings demonstrate our method's effectiveness at separating novel and old class samples continuously. We will release our code upon acceptance. 
\end{abstract}

\vspace{-5mm}
\section{Introduction and Related Work}
\vspace{-2mm}
Deployed AI models frequently encounter dynamic and evolving data distributions, where continuous model adaptation is paramount to safeguard performance. 
Reliable novelty detection is a key capability for adaptive AI. Novelty Detection will inform the model if there is new data and if so, which samples are novel and need to be learnt from.
However, until now, novelty detection and continual adaptation have been tackled separately within different sub-fields of the AI scientific literature. Most research in continual learning (CL) \cite{kirkpatrick2017overcoming,wen2021beneficial,cheung2019superposition,rebuffi2017icarl,rios2020lifelong,buzzega2021rethinking} relies on fully labeled data, despite the significant costs and impracticality of data labeling in real-world scenarios \cite{parisi2019continual}. While there are some unsupervised CL solutions \cite{ordisco,Boschini2022,bagus2022supervised}, they often rely on an unrealistic assumption: that for each new task and its incoming data, past classes do not appear alongside newly introduced classes, thereby eliminating the need for novelty detection. Removing this oracle assumption results in severe performance degradation due to overconfidence in erroneous predictions \cite{rios2022incdfm}: novel classes' samples may be incorrectly predicted to old classes, especially at task transition onset where the continual decision boundaries are still immature. 
Meanwhile, solutions for novelty or out-of-distribution (OOD) detection \cite{hendrycks2016baseline, liang2018enhancing, ndiour2020probabilistic, lee2018simple, haoqi2022vim, ren2019likelihood} have primarily been designed and evaluated using a single, fixed split of old versus novel classes, rather than on continual splits. Additionally, conventional OOD models often lack the ability to continuously integrate and learn from newly detected data. When these models are forced to update, they can suffer from continual error propagation \cite{rios2022incdfm}: incorrect novelty predictions during the detection stage lead to incorrect parameter learning during the update stage, progressively degrading the overall system performance.
The recently proposed incDFM \cite{rios2022incdfm} offers an innovative solution to continual novel class detection (CND). However, incDFM was designed for the simplistic scenario where only one novel class is introduced per task. This strong assumption allows incDFM to treat all samples flagged as novel
as members of the single new class, enabling trivial pseudo-labeling for continual update. Due to this unrealistic one-class assumption, incDFM cannot be considered fully unsupervised. In more complex cases with multiple novel classes, incDFM fails to function effectively since it cannot distinguish between different novel classes. In effect, all samples from those multiple novel classes are erroneously assigned to a single new class. This multi-class "collapse" results in a poor estimate of the OOD/novelty distribution and consequently, poor performance. Generalizing to the scenario of continuous multi-class novelties is challenging, necessitating the creation of entirely new algorithmic components. 
\textbf{Our contribution is as follows:} We propose CONCLAD (COntinuous Novel CLAss Detector), an iterative multi-class uncertainty estimation algorithm designed for generalized Continual Novelty Detection. We utilize the uncertainty scores to select (a) a very small fraction (0.3\% - 1.25\%) of samples from the unlabeled pool for supervision and (b) a suitable subset of the remaining unlabeled samples for automatic (unsupervised) pseudo-labeling. Through experimentation on various continual tasks and datasets, we demonstrate that CONCLAD excels in continually identifying the presence of (up to multiple) novel classes and accurately separating novel class samples from old ones.
\vspace{-4mm}
\section{Our Method}
\vspace{-3mm}
\textbf{2.1. Problem Setting:} 
Consider a continual agent $A(x,t)$ which needs to learn/adapt from a set of continual tasks. At each task $t$, $A(x,t)$ is presented with an initially unlabeled set of samples $U(t)$ \ref{eq:setup} which consists in a mixture of unseen samples of its old/learnt classes $U_{old}(t)$ and unseen samples of new (novel) classes $U_{new}(t)$: 
\vspace{-4mm}
\begin{gather}
U(t) = U_{old}(t) \cup U_{new}(t), \text{where } 
U_{old}(t)  = \{x | x \thicksim \bigcup_{k=1}^{t-1} D_k\}, U_{new}(t) = \{x | x \thicksim D_t\},
\label{eq:setup}
\end{gather}
Here $D_t$ comprises samples from the set of new classes $C^t_{new}$ introduced at task $t$, while $\bigcup_{k=1}^{t-1} D_k$ are samples belonging to all the old classes $C_{old}^{t}$ that have been learned up to and including task $t-1$. Samples in $U_{old}(t)$ are ``unseen'', meaning they were never used, neither in the initial training nor during prior tasks' learning. Note that addressing data drifts in $U_{old}$ is beyond the scope of this work.




\textbf{2.2. Our solution:} We introduce a continual novelty detector $N(x,t)$, operating alongside the continual agent, whose goal is to produce a reliable estimate of novel samples $\widehat{U}_{new}(t)$
while simultaneously estimating their respective novel-class labels. Simply performing a binary distinction between novel-class and old-class samples (as in incDFM\cite{rios2022incdfm}) leads to poor results in novel multi-class settings. Moreover, the dependence on task index $t$ in $N(x,t)$ indicates that the novelty detector itself has to be continually updated so that novel classes at $t$ are not considered novel at $t+1$. 
To obtain novel-class labels in $\widehat{U}_{new}(t)$, one can either used unsupervised clustering methods \cite{ren2024deep}, or active supervision (i.e. labeling by an expert) \cite{nguyen2022measure,gal2017deep,yoo2019learning}. Here, we share initial results using active supervision for a tiny fraction of $U(t)$ (0.3\% - 2.5\%), along with pseudo-labeling of confidently identified novel samples in $\widehat{U}_{new}(t)$. 
For all these tasks -- novelty detection, sample selection for active labeling, and for pseudo-labeling -- $N(x, t)$ relies foundationally on a novel, iterative multi-class uncertainty estimation method \ref{eq:uncertainty-uncanny} defined and explained in the next sections.

\textbf{2.2.1. Building block of CONCLAD's uncertainty formulation $S(i)$:}
CONCLAD's  uncertainty estimation \ref{eq:uncertainty-uncanny} uses the \textit{feature reconstruction error} (FRE) \cite{ndiour2020probabilistic}, which is effective in novelty estimation for the closed-world and the single-class increment CL \cite{rios2022incdfm}. FRE involves learning a PCA transform $\mathcal{T}_m$ and its inverse $\mathcal{T}_m^{\dagger}$ for each class $m$. A test feature $u=g(x)$ is transformed by $\mathcal{T}_m$ and re-projected back using $\mathcal{T}_m^{\dagger}$, with FRE calculated as the $\ell_2$ norm of the difference between the original and reconstructed vectors. High FRE scores indicate samples that don't belong to class $m$. In the simplified single-class increment CL \cite{rios2022incdfm}, a single PCA transform is used for all ID data.

\textbf{2.2.2. Step by Step Novelty Detection:}
Prior to deployment (task $t=0$), we assume that an agent $A(x,t=0)$ has been trained to classify among a fixed set of pre-deployment classes $C_{new}^0$. Accordingly, CONCLAD's novelty detector $N(x,t=0)$ has been trained to recognize those classes as learnt/old by having computed FRE transforms for those classes, $\mathcal{T}_m, \forall m \in C_{new}^0$. For a given future task $t>0$, as unlabeled data arrives, $N^{(i)}(x,t)$ follows an iterative procedure (indexed by an inner-loop index, $i$, which is distinct from outer-loop task-index $t$) to learn to detect if/what novelties are present. 
At the first inner iteration $i=0$,
initial supervision querying is performed by picking 
samples (subject to labeling budget) with high uncertainty scores w.r.t old classes defined as $S^0(u) \triangleq \min_{j\in C_{old}^t} FRE_j^0(u)$. $b_0$ is sampled uniformly among samples with $S^0(u)>\text{mean}(S^0(u))$. 
At this point, novel classes can be identified (denoted by $|C_{new}^t|$ in section 2.1, assuming $|C_{new}^t|> 0$) and those few labeled samples are used to initialize parameters of $N^{(0)}(x,t)$: (1) Train a single layer perceptron, $N_{pl}^{(i)}(x,t)$ to learn an imperfect initial mapping to the $|C_{new}^t|$ novel classes. This layer, which performs pseudo-labeling (pl), contains output nodes only w.r.t novel classes.
(2) compute rough estimates of per-novel-class PCA transforms $\{\mathcal{T}^{t,0}_m\}, m\in C_{new}^t$. Note that it's possible that not all true novel classes are found in this initial iteration and may be found in subsequent ones. For subsequent iterations $i>0$, given an unlabeled sample $x \in U(t)$, $N_{pl}^{(i)}(x,t)$ predicts a pseudo-label $m, m \in C_{new}^t$ which then routes the selection of the corresponding PCA transform $\mathcal{T}^{t,i-1}_m$ resulting in the $i^{text{th}}$ iteration's uncertainty score $S^i(x)$ \ref{eq:uncertainty-uncanny}:
\vspace{-1mm}
\begin{equation}
\label{eq:uncertainty-uncanny}
    S^i(u) = \min_{j\in C_{old}^t}\frac{FRE_j^0(x)}{FRE^{i-1}_m(x)}; i>0, m=N_{pl}^{(i)}(x,t) \in C_t^{new}\\
\end{equation}
$S^i(x)$ can be used to robustly categorize samples in $U(t)$ as: \textbf{(1) Novel with high-confidence:} These are samples with the highest score values (high numerator relative to the denominator). A high value of numerator implies large distance from previously seen classes $C_{old}^t$, while a low value of the denominator implies low distance from novel class $m$. Such a sample likely belongs to $U_{new}(t)$ and is a strong candidate to be pseudo-labeled.
From these, we select the topmost most confident $\alpha$ percent to pseudo-label. \textbf{(2) Old-class with high-confidence:} lowest score values corresponding to low numerator (low distance w.r.t $C_{old}^{t-1}$) and high denominator value (high-distance from the predicted novel class $m$). Such a sample likely belongs to $U_{old}(t)$, i.e. to an old class that has already been learned; \textbf{(3) Ambiguous:} Samples for which the score is neither definitively high nor definitively low. These could be old-class samples having relatively high scores, or new-class samples having relatively low scores. Owing to this ambiguity, a clear determination cannot be made. Hence, these samples are excellent candidates for active querying to minimize novelty detection uncertainty.
At each inner-loop iteration, accumulated active and pseudo-labeled samples are used to re-update $N^{(i+1)}(x,t)$'s parameters (pseudo-labeler $N_{ps}^{(i)}(x,t)$, and FRE transforms $\mathcal{T}^{t,i-1}_m$). At the end of the inner-loop, all accumulated active and pseudo labeled samples are used to compute final PCA transforms $\{\mathcal{T}^{t}_m\}$ for $m \in C_t^{new}$ to permanently update the novelty detector $N(x,t)$ so those classes are not flagged as novel subsequently. Note that the pseudo-labeler, since it maps only to a given tasks detected novel classes, will be re-initialized at another tasks' onset. Further methodology details, including inner-loop stopping criteria and ambiguity formulation, can be found in appendix sections.

\vspace{-3mm}
\section{Experiments}
\vspace{-2mm}
\textbf{3.1. Setup: }
We evaluate on 4 datasets: Imagenet21K-OOD (Im21K-OOD) \cite{ridnik2021imagenet21k}, Eurosat \cite{helber2019eurosat}, iNaturalist-Plants-20 (Plants) \cite{vanhorn2018inaturalist} and Cifar100-superclasses \cite{cifar100}, all of which were constructed to have no class overlap with
Imagenet1K with the exception of Cifar100. 
Results for Cifar100 are included to enable direct comparison with baseline method incDFM \cite{rios2022incdfm}. We compare CONCLAD to: (1) incDFM \cite{rios2022incdfm}, which first introduced an updatable continual novelty detector, albeit exclusively for single class novelties (see section 1); (2) DFM \cite{ndiour2022subspace}, originally proposed for static novelty detection. We also include semi-supervised CND baselines: (3) Experience-Replay "ER" \cite{rolnick2019experience, buzzega2021rethinking} uses entropy as a measure of novelty similar to \cite{aljundi2022continual} and also to select active labels; (3) PseudoER \cite{pseudoDER}, same as ER, but iteratively pseudo-labels the most confident samples akin to CONCLAD. Other baselines are constructed (Fig 2 right table) from removing elements of CONCLAD such as the iterativeness (i.e. doing AL/Pseudo-labeling in one shot), etc. Implementations for CONCLAD and baselines: All use a large/foundation frozen feature extractor, e.g. ResNet50 \cite{he2016deep} pre-trained on ImageNet1K via SwAV \cite{caron2020unsupervised} or ViTs16 \cite{alexey2020image} pre-trained on Imagenet1K via DINO \cite{caron2021emerging}. CONCLAD's $A_s^{cl}$ (pseudo-labeling head) is a fully connected layer. Baselines ER, PseudoER's long-term classification head is a perceptron of size 4096. For ER and PseudoER we use a fixed replay buffer size containing pre-logit deep-embeddings and labels/pseudo-labels. We set the maximum buffer size to 5000 (2500 for Eurosat). At each incoming unlabeled pool, we fix a mixing ratio of 2:1 of old to new classes per task, with old classes drawn from a holdout set (0.35\% of each dataset). For evaluation on the independent test set, we sample old and new classes with the same 2:1 proportion. Note that old classes act as distractors from the point of view of novelty detection. We set pseudo-labeling selection to $\alpha=20\%$ of samples predicted as novel (appendix 4.1.1). For experiments not purposely varying the tiny supervision budget, we fix a labeling budget of 1.25\% for Places, Plants and 0.625\% for Eurosat, Im21K-OOD, as guided by Fig 1 \textit{center} which varies the AL budget from 0.625\% to 5\%.

\textbf{3.2. Results: }
We measure continual novelty detection performance with the common "Area Under the Receiver-Operating-Curve" (AUROC) metric. Note that, for fair evaluation, we measure CND on an independent test set with the same ratio of old to new class samples at each task. Fig. 1 (left) displays CND performance (AUROC) over all continual tasks (time) in the case of multi-class novelties per task (5 class increments for Im21K-OOD and 2 class increment for Eurosat). Additionally, figure X (center) shows the sensitivity of CONCLAD and other actively-supervised baselines (ER-entropy, PseudoER-entropy, section 3.1) when varying the tiny supervision budget (tested for a range of 0.32\% to 5\% of the unlabeled train data at each task). Fig. 1 (right) shows the effect of varying the novel class increment per task as measured by the AUROC score averaged over all continual tasks (with that given increment). Some interesting highlights: (1) we can see in Fig 1 (right) that the compared approach incDFM \cite{rios2022incdfm} performs reasonably well for the increment of only one novel class per task, for which it was originally proposed and tested by the authors. However, when the class increment increases, this method degrades in performance because it groups multiple novel classes with no distinction, which hurts detection. (2) PseudoER consistently under-performs ER because it is unable to produce high confidence pseudo-labels to be used in training and this in turn degrades its performance - this highlights the importance of our uncertainty metric \ref{eq:uncertainty-uncanny} in measuring pseudo-label confidence. 
(3) It is evident in the above plots that even with tiny supervision budgets (e.g. 0.32\%-1.25\%), CONCLAD consistently outperforms the competing methods by a large margin over the several experimental variations.

\vspace{-3mm}
\begin{figure}[h]
\centering
\includegraphics[width=14.2cm]{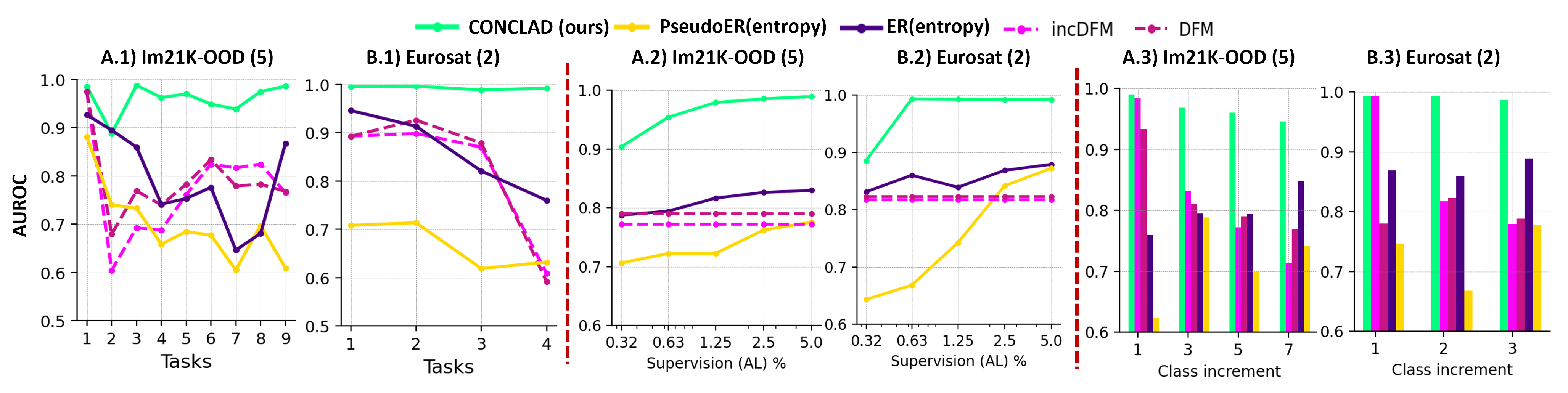}
\vspace{-6mm}
\caption{(\textit{Left} A.1,B.1) Continual Novelty Detection performance measured by AUROC at each task. The number of novel classes introduced per task is in parenthesis. Overall, CONCLAD (green) significantly over-performs baselines; (\textit{Center} A.2,B.2) Results varying the supervision budget; (\textit{Right} A.3,B.3) Results varying Novel Class Increment per task. For (left,right) Supervision budget is 0.625\% for CONCLAD, ER, PseudoER. Equivalent plots for Cifar100, Plants in appendix 4.3.}
\label{fig:plots}
\vspace{-7mm}
\end{figure}

\vspace{-2mm}
\begin{figure}[h]
    \scriptsize
    \centering
    \captionsetup[sub]{labelformat=empty,   
                   position = top}
    \subcaptionbox{}{
        \addtolength{\tabcolsep}{-0.43em}
        \begin{tabular}
        {c|cc|cc|cc|cc} 
        \toprule
        {} & \multicolumn{2}{c|}{Im21K} & \multicolumn{2}{c|}{Plants} & \multicolumn{2}{c|}{Eurosat} & \multicolumn{2}{c}{Cifar100}\\ 
        {} & R50 & ViT & R50 & ViT & R50 & ViT & R50 & ViT\\
        \midrule 
        CONCLAD(ours) &\textbf{96.0} &  \textbf{88.0}&  \textbf{73.6}&  \textbf{58.8} & \textbf{99.3}& \textbf{82.3}& \textbf{80.6}& \textbf{83.3}\\
        incDFM & 77.3&  76.0&  68.7& 58.2 & 81.8 & 74.9& 66.3& 66.3\\
        DFM & 79.0 &  75.4&  67.4&  54.9& 82.2& 74.9& 62.2& 66.3\\
        ER-Entropy & 79.4&  52.8& 61.4 & 52.4 & 86.0& 46.6& 64.1& 55.4\\
        PseudoER-Entropy & 69.9&  54.2& 60.6 &  52.5& 66.8& 46.5& 59.5& 52.6\\
        \midrule 
        \end{tabular}
    }
    \subcaptionbox{}{
        \raisebox{0.05\height}{
        \begin{tabular}[b]{p{1.2cm}p{0.5cm}p{0.5cm}p{0.5cm}p{0.5cm}}
        \toprule
        {Variations (R50)} & {Im21K} & {Plants} & {Eurosat} & {Cifar100}\\ 
          \midrule
          Default &  \textbf{96.0} &  \textbf{73.6} &  \textbf{99.3} &  80.6\\
          \midrule
          Sup-Top & 89.5 & 71.0 & 97.6 & \textbf{81.9}\\
          Sup-Rand&  88.7&  65.9&  98.9& 80.4\\
          No-Iters &  89.5& 68.2 & 80.5 & 66.4\\
          No-Pseudo &  77.2&  65.4& 75.6 & 64.1\\
          \hline
        \end{tabular}
        }
        }
    \vspace{-2mm}
    \caption{(Left) Continual Novelty Detection measured by AUROC; (Right) Ablations of CONCLAD. Supervision budget is 0.625\% for Im21K, Eurosat and 1.25\% for Plants, Cifar100}
    \label{fig:tables}
    \vspace{-3mm}
\end{figure}

Fig 2 table (left) shows average AUROC results over all tasks for all 4 datasets and with two different feature extraction backbones (Vision Transformer "ViT" and Resnet50 "R50" described in section 3.1). In sum, similar conclusions can be reached here: CONCLAD significantly overperforms baselines over all the tested settings. Additionally, Fig 2 table (right) shows results for different ablations of CONCLAD: (\textit{No-Pseudo}) Removing Pseudo-labeling from $S(i)$, i.e. computing per-novel class PCAs only with ground-truth label assignments obtained with the tiny labeling budget; (\textit{Sup-Random}) Using random sampling to query ground truth labels with the same tiny budget; \textit{Sup-Top} queries samples with highest uncertainty scores (i.e. most-confidently novel samples) for ground-truth labeling rather than ambiguous samples;(\textit{No-Iters}) CONCLAD in oneshot. Use all supervision budget upfront and then pseudo-label in one-shot. The ablation results highlight the importance of minimizing error propagation via our method's iterativeness since No-Iters results in an average 11.2\% decrease in performance. Similarly, we show that pseudo-labeling among the multiple novel classes detected is fundamental to performance given the AL budget's tiny size: No-Pseudo results in 16.8\% average decrease. Finally, other active labeling strategies (i.e. Sup-Top) or lack-thereof (Sup-Rand) also decrease performance by 2.4\% and 3.9\% respectively, underscoring the informativeness of querying ambiguous samples for AL with the goal of continual novelty detection, refer to section 2.2.2. 

\textbf{Key Takeaways:} 
In this work, we presented CONCLAD, a solution to the still under-explored problem of continual novelty detection (CND). Our method enables CND in the generalized setting of novelties containing up to multiple novel classes. To achieve this, CONCLAD includes a foundationally novel iterative multi-class uncertainty estimation procedure capable of effectively modelling the distribution of multiclass novelties, only with a tiny supervision budget. By minimizing the number of samples falsely flagged as novel or overlooked as old, we ensure minimal continual error propagation.
Overall, CONCLAD outperforms baselines over multiple large-scale datasets and experimental variations. Yet, several challenges remain for CND, which we hope to address in future work. One nontrivial example is how to detect both novel classes and distribution shifts of old classes (e.g. noise, illumination, etc) together, with minimal to no supervision.




{\small
\printbibliography
}

\input{appendix.tex}



\end{document}

%% file: appendix.tex
\section{Appendix}
\subsection{Methodology Details}

\subsubsection{Thresholds for Stopping the inner-loop} The inner-loop is guided by two simple thresholds: (1) Threshold $T_{inner}$ "roughly" estimates if there are any possible novel-class samples in the unlabeled task input data pool and is controlled by a single hyper-parameter, the number of standard deviations above the mean of an in-distribution validation set (2 STDs in our experiments). If no samples are found to be above $T_{inner}$, we reach the stopping criterion for our iterations. Our in-distribution validation set is conventionally defined to include a portion (0.1\%) of the previous tasks' $k=1:t-1$ novelty predictions that were held-out at previous tasks, i.e. not used to update $N(x,t)$ parameters. Importantly, the same in-distribution validation set is used for all compared baselines in our results section, as is common practice in the OOD/novelty-detection literature \cite{rios2022incdfm, hsu2020generalized}; (2) Finally, Threshold $\alpha$ tunes pseudo-labeling selection and is set to $\alpha = 20\%$ highest $S^i(u)$ scores (most confident) from the test samples found above $T_{inner}$. These two thresholds are not highly sensitive.

\subsubsection{How to define Ambiguity} 
CONCLAD seeks to minimize novelty-detection uncertainty and model multiclass-novelties by selecting the most novel-vs-old ambiguous samples at each inner-loop iteration, i.e. scores $S^i(u)$ which are neither too high or too low. Our mathematical formulation uses the threshold $T_{inner}$ defined in the previous section: we formulate ambiguousness as the inverse squared distance $\frac{1}{\|S^i(u)-T_{inner}\|^2}$ of scores to $T_{inner}$. Intuitively, this formula favors selecting samples that cannot be unambiguously predicted as either old or new since $T_{inner}$ represents this rough decision boundary. Active selection is stopped when the tiny labelling budget is exhausted. The only exception to this Ambiguity formulation is at the first iteration $i=0$ where we select homogeneously from samples above $T_{inner}$. This is the case because at $i=0$ only old classes are used to compute the score function, $S^0(u) = \min_{j\in C_{old}^t} FRE_j^0(u)$ and so ambiguity cannot be defined in the same way as for the remainder of iterations.

\subsubsection{Measuring per-class uncertainty in CONCLAD's formulation}
CONCLAD is agnostic to the elemental uncertainty metric used in its uncertainty scoring function ($S^i(u)$ Eq. 2 in section 2) as long as it can reliably estimate uncertainty w.r.t each novel class or old class. However, this is not an easy feat since many existing static uncertainty quantification approaches are not fully reliable \cite{ndiour2020probabilistic,rios2022incdfm}. As discussed in the main text, CONCLAD currently leverages the \textit{feature reconstruction error} (FRE) metric introduced in \cite{ndiour2020probabilistic} to build Eq 2. For each in-distribution class, FRE learns a PCA (principal component analysis) transform $\{\mathcal{T}_m\}$ that maps high-dimensional features $u$ from a pre-trained deep-neural-network backbone $g(x)$ onto lower-dimensional subspaces. During inference, a test-feature $u=g(x)$ is first transformed into a lower-dimensional subspace by applying $\mathcal{T}_m$ and then re-projected back into the original higher dimensional space via the inverse $\mathcal{T}_m^{\dagger}$. 
The FRE measure is calculated as the  $\ell_2$ norm of the difference between the original and reconstructed vectors:
\begin{equation}
\label{eq:FRE}
    FRE_m(u) = \|f(\mb{x})-(\mathcal{T}_m^{\dagger} \circ \mathcal{T}_m)u\|_2.
\end{equation}
Intuitively, $FRE_m$ measures the distance of a test-feature to the distribution of features from class $m$. If a sample does not belong to the same distribution as that $m$th class, it will usually result in a large reconstruction score $FRE_m$. FRE is particularly well suited for the continual setting since for each new class discovered at test-time, an additional principle component analysis (PCA) transform can be trained without disturbing the ones learnt for previous classes.

\subsection{Experimental Methodology Details}
\subsubsection{Implementation Details for CONCLAD and Baselines} 
$N(x,t)$ operates on top of a large-scale/foundation models as feature extraction backbones, kept frozen throughout CONCLAD and baselines' training: (1) Most results use ResNet50 \cite{he2016deep} unsupervisedly pre-trained on ImageNet1K via SwAV \cite{caron2020unsupervised}. We extract features from the pre-logit AvgPool layer of size 2048 as deep-embeddings. We also experimented with other feature extraction points \cite{ndiour2020probabilistic} but those under-performed w.r.t the pre-logit layer. (2) We also show results using ViTs16 \cite{alexey2020image} pretrained on Imagenet1K via DINO \cite{caron2021emerging}. For ViTs16 we tried several extraction points, e.g. head, last norm later, different transformer block outputs with different pool factors (e.g. 2,4). Best results were obtained with the norm layer. Note that learning on frozen deep features is commonplace in vision CL and domain-adaptation fields \cite{rios2020lifelong,rios2022incdfm,evci2022head2toe}. It is theoretically based on the principle that low-level visual features from a large-scale/foundation frozen model are task nonspecific and do not need to be constantly re-learned. Rather, learning may happen upstream by utilizing the extracted deep features (at the last or inner-layers, or a combination thereof - an active research area) \cite{petrov2005dynamics,dhillon2019baseline,evci2022head2toe}. CONCLAD's $N(x,t)$ fully-connected pseudo-labeling layer is trained with ADAM \cite{kingma2014adam}, learning rate of 0.001, mini-batch of 10 and an average of 5 epochs at each inner-loop. We experimented with other possibilities of pseudo-labeler such as a 1-layer perceptron but obtained marginal performance gain. Baselines' ER and PseudoER long term classification head are implemented as a one layer perceptron of size 4096 (also tested variations with marginal variations in results). The ER/PseudoER replay buffer is set to a size 5000 deep-embeddings for Plants \cite{vanhorn2018inaturalist}, Imagenet21K-OOD \cite{ridnik2021imagenet21k} and 2500 for eurosat and cifar100. We use a fixed-size memory buffer $B_t$ with the same building strategy and training loss as in \cite{rios2018closed}: a buffer of fixed size and prioritizing homogeneous distribution among classes. That is, an equivalent number of samples of each class are removed if room is required for new classes and the buffer is full. Equal weight is given to old and new classes during ER. Lastly, baselines incDFM \cite{rios2022incdfm} and DFM \cite{ndiour2020probabilistic} were trained using same hyper-parameters proposed by the authors and their open-source code.



\subsubsection{Datasets:} 
Since the employed large/foundation feature extractor were pretrained on Imagenet1K,
we evaluate CONCLAD on datasets that either do not contain class overlap with Imagenet1K (out-of-distribution w.r.t Imagenet1K \cite{2aa757143d6f46e2aba527d9e1a26aa5}), or curated them by excluding any overlapping classes. The exception is cifar100, which was included due to it being a very popular and widespread dataset, also used in incDFM \cite{rios2022incdfm}. 
\begin{enumerate}
    \item \emph{Imagenet21K-OOD (Im21K-OOD) \cite{ridnik2021imagenet21k}}: We curated a subset of Imagenet21K containing the top-most populous 50 classes and that do not overlap with the classes present in Imagenet1K. We use a random set of 500 samples from each of the 50 classes. Because Imagenet21K is a superset of Imagenet1K, by excluding any overlapping class we guarantee orthogonality in our curated subset. We will release the full list of images chosen in this curation for reproducibility. 

    \item \emph{iNaturalist-Plants-20 (Plants) \cite{vanhorn2018inaturalist}}: is a curated subset containing images from 20 OOD plant species, sourced from the iNaturalist project \cite{vanhorn2018inaturalist}. A super-set (larger) version of this subset was originally proposed by \cite{huang2021mos} and has since been frequently used as test OOD dataset with respect to Imagenet1K \cite{ming2022delving}. Note that we use only 20 classes instead of the original 110 in the \cite{huang2021mos} super-set since we remove classes with sample count below 140.

    \item \emph{Eurosat \cite{helber2019eurosat}}: An RGB dataset of 10 classes and 27K images of Sentinel-2 satellite images, which is also orthogonal to Imagenet1K.

    \item \emph{Cifar100-Superclasses (Cifar100) \cite{cifar100}}: We use the super-label granularity of Cifar100 dataset. This totals 20 labels (super) and 50K images. While Cifar100  is not orthogonal to Imagenet1K, we decided to showcase its results since it is a widespread dataset in CL. 
\end{enumerate}

\subsubsection{Baselines} 
For continual novelty detection (CND), we include unsupervised baselines that also utilize FRE-based uncertainty measures: DFM \cite{ndiour2022subspace} and incDFM \cite{rios2022incdfm}. The latter, incDFM \cite{rios2022incdfm}, was the first to develop an updatable continual novelty detector for CND, albeit exclusively tested for the trivialized case of single class novelties only, see discussion in main paper section 1. Alternatively, DFM originally introduced the FRE measure \ref{eq:FRE} for static novelty detection. In the case of incDFM, their proposed scoring function after training/update could be directly used to compute novelty detection on a test set, in the continual setting. We use the author's official implementation of incDFM to generate results. For DFM, we adapted the method to the continual setting by storing one PCA transform $\mathcal{T}_j$ per task trained from all data predicted as novel at the previous task. The scoring function $S_{DFM}^{t}$ for DFM is defined in equation \ref{eq:uncertainty-DFM}, with $T_{old}^t$ representing the count of how many past tasks with novelty(ies) have previously occured at time/task $t$.
\begin{equation}
\label{eq:uncertainty-DFM}
    S_{DFM}^{t}(u) = \min_{j\in T_{old}^t}FRE_j(u)
\end{equation}
We also include semi-supervised baselines, with the same tiny supervision budget: (2) ER \cite{rolnick2019experience, buzzega2021rethinking}, originally proposed for supervised CL is adapted to only use actively labeled samples (as embeddings) for replay; (3) We also adapt PseudoER \cite{pseudoDER} similar to ER but further incorporating pseudo-labeling of high confidence unlabeled samples for training. In both ER and PseudoER, we utilize the cumulative classification entropy as an uncertainty score to actively-label and Pseudo-Label (PseudoER). Similar to CONCLAD, we actively label ''ambiguous'' samples according to the same formula as outlined in appendix 4.1.2 for superior results, then sampling according to the TOP heuristic (see section 3 discussion). We also tested with other common uncertainty metrics such as margin \cite{ren2021survey} but with inferior results. 

\subsection{Additional Results}
\begin{figure}[h]
\centering
\includegraphics[width=14cm]{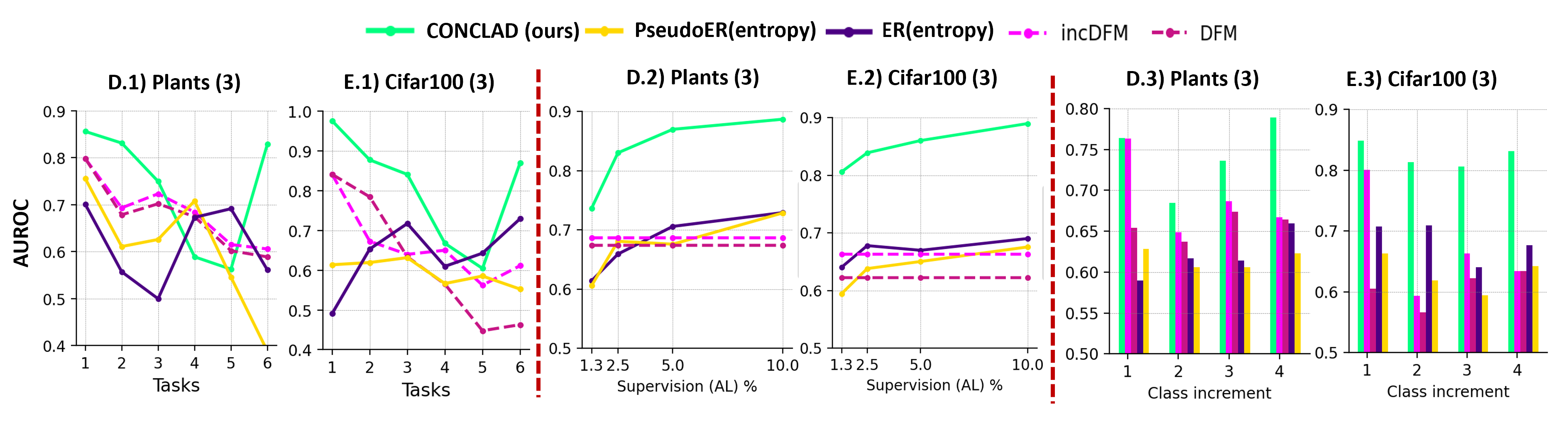}
\caption{Results for Plants and Cifar100; (\textit{Left} D.1,E.1) Continual Novelty Detection performance measured by AUROC at each task. The number of novel classes introduced per task is in parenthesis.(\textit{Center} D.2,E.2) Results varying the supervision budget; (\textit{Right} D.3,E.3) Results varying Novel Class Increment per task. For (left,right) Supervision budget is 1.25\% for CONCLAD, ER, PseudoER. Overall, CONCLAD (green) significantly over-performs baselines}
\label{fig:cifar}
\end{figure}